\documentclass[conference]{IEEEtran}
\IEEEoverridecommandlockouts
\usepackage{cite}
\usepackage{amsmath,amssymb,amsfonts}
\usepackage{algorithmic}
\usepackage{graphicx}
\usepackage{textcomp}
\usepackage{xcolor}

\usepackage{graphics} 
\usepackage{epsfig}
\usepackage{booktabs} 
\usepackage{multirow}
\usepackage{caption}
\usepackage{float} 
\usepackage{subcaption}
\def\BibTeX{{\rm B\kern-.05em{\sc i\kern-.025em b}\kern-.08em
    T\kern-.1667em\lower.7ex\hbox{E}\kern-.125emX}}
\begin{document}

\title{Knowledge Augmented Entity and Relation Extraction for Legal Documents with Hypergraph Neural Network}

\author{\IEEEauthorblockN{1\textsuperscript{st} Binglin Wu}
\IEEEauthorblockA{\textit{School of Economics and Management} \\
\textit{Dalian University of Technology}\\
Dalian, China \\
binglin@mail.dlut.edu.cn}
\and
\IEEEauthorblockN{2\textsuperscript{nd} Xianneng Li\IEEEauthorrefmark{1}
\thanks{\IEEEauthorrefmark{1} Corresponding author}}
\IEEEauthorblockA{\textit{School of Economics and Management} \\
\textit{Dalian University of Technology}\\
Dalian, China \\
xianneng@dlut.edu.cn}
}

\maketitle

\begin{abstract}
With the continuous progress of digitization in Chinese judicial institutions, a substantial amount of electronic legal document information has been accumulated. To unlock its potential value, entity and relation extraction for legal documents has emerged as a crucial task. However, existing methods often lack domain-specific knowledge and fail to account for the unique characteristics of the judicial domain. In this paper, we propose an entity and relation extraction algorithm based on hypergraph neural network (Legal-KAHRE) for drug-related judgment documents. Firstly, we design a candidate span generator based on neighbor-oriented packing strategy and biaffine mechanism, which identifies spans likely to contain entities. Secondly, we construct a legal dictionary with judicial domain knowledge and integrate it into text encoding representation using multi-head attention. Additionally, we incorporate domain-specific cases like ``joint crimes" and ``combined punishment for multiple crimes" into the hypergraph structure design. Finally, we employ a hypergraph neural network for higher-order inference via message passing. Experimental results on the \textbf{CAIL2022} information extraction dataset demonstrate that our method significantly outperforms existing baseline models.
\end{abstract}


\section{Introduction}
In recent years, the digitalization of judicial institutions in China has advanced significantly, resulting in the accumulation of a substantial amount of electronic legal document data, such as judgment documents, laws and regulations, and contracts. However, due to the diversity of judicial text data structures and variations in the writing habits of judicial personnel, simple rule-based matching yields an extremely low information coverage rate, causing the omission of significant information. Therefore, the extraction of key information from complex legal texts has become a critical research topic in legal artificial intelligence. In existing research, knowledge bases are often constructed from legal texts using information extraction technology to serve specific downstream tasks such as legal question answering \cite{bach2017question}, similar case retrieval \cite{van2017concept,tran2019building}, and legal judgment prediction \cite{hu2018few,chen2019charge}. Therefore, enhancing the effectiveness of information extraction can significantly benefit downstream tasks. In addition, the elements extracted from legal documents are grounded in legal principles and can provide explanations for legal applications \cite{zhong2020does}.

Entity and relation extraction, as a crucial subtask within the field of information extraction, aims to extract both predefined entity information and subject-relation-object triples from unstructured texts. In the judicial domain, the extracted entity relations typically constitute the key information of criminal cases, including the names of suspects, specific criminal behaviors, and crime locations. This process transforms unstructured information within legal documents into structured triples, facilitating the construction of judicial knowledge graphs \cite{li2019survey}. Existing research can be divided into two main types: the pipeline approach and the joint approach. The pipeline approach first performs named entity recognition and then proceeds to relation extraction \cite{zheng2016neural,chen2022pattern}. However, this method disregards the interactions between entities and relations, and suffers from error propagation. In contrast, the joint approach, which is a sequence-to-sequence method, conducts unified modeling of named entity recognition and relation extraction \cite{liang2022sequence,gao2023ergm}.

\begin{figure*}
      \centering
      \includegraphics[width=0.98\linewidth]{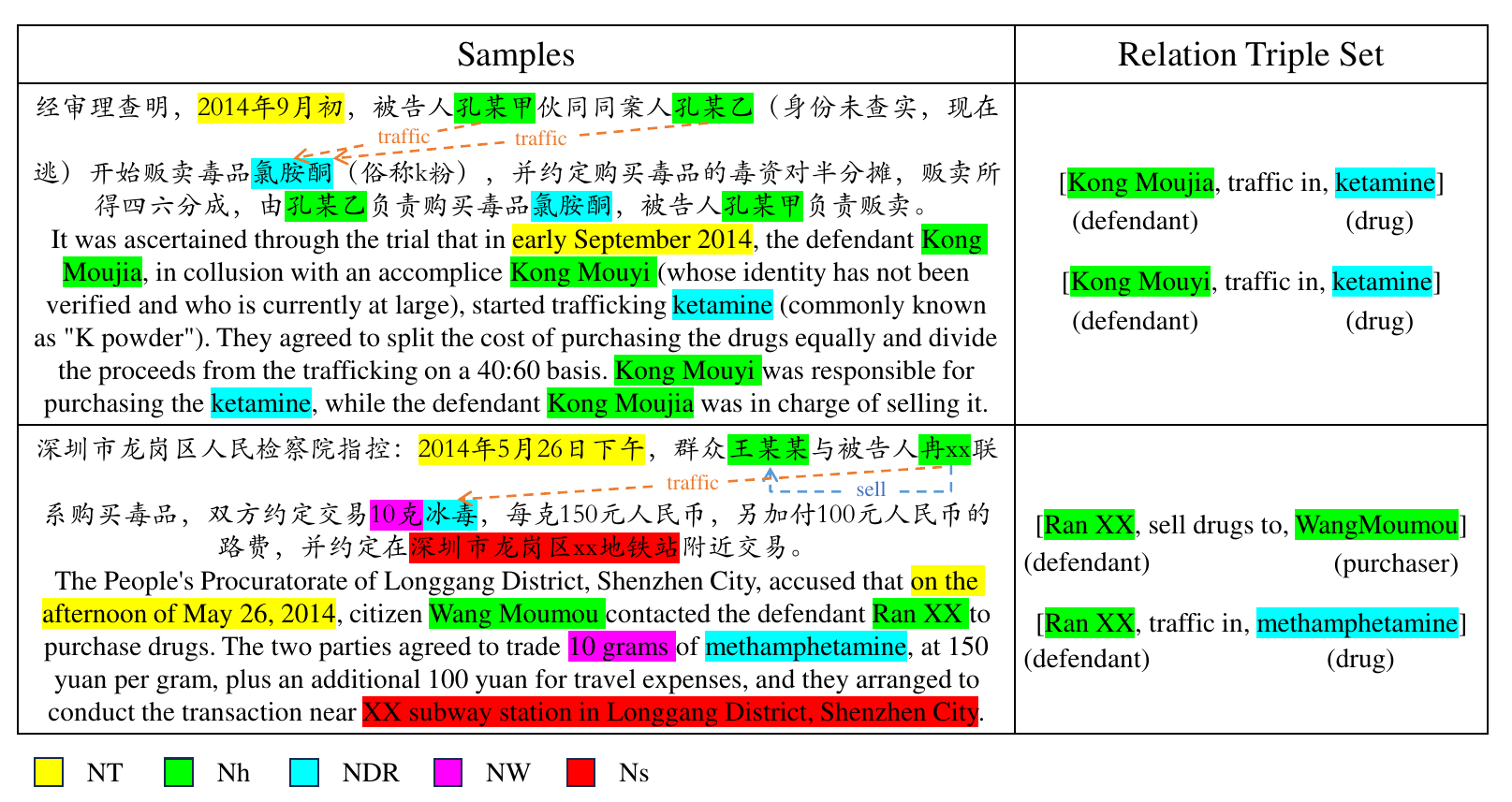}
      \caption{Examples of special situations in the judicial field}
      \label{fig1}
\end{figure*}

However, the judicial domain presents unique challenges for entity and relation extraction. Judicial documents are often complex texts that contain a large number of professional terms and structured information, making it difficult for models to understand. The high stakes in the judiciary require extremely high accuracy in the extracted key information, with little tolerance for errors. Additionally, the judicial field encompasses specific situations, such as joint crimes and combined punishment for multiple crimes, as shown in Fig.~\ref{fig1}. In Fig.~\ref{fig1}, Kong Moujia and Kong Mouyi jointly trafficked ketamine, and Ran XX not only sold drugs to Wang Moumou but also trafficked methamphetamine. Moreover, these situations often appear as independent relation triples in the dataset, and existing models have not effectively utilized this information.

To address the above-mentioned challenges, we propose a knowledge augmented entity and relation extraction method based on hypergraph neural networks, designed for extracting entities and relations in legal documents related to drug cases. Our model consists of four components: a candidate span generator, a legal feature-augmented encoder, a node initialization module, and a hypergraph neural network module. The candidate span generator, pre-trained independently of the other modules, identifies spans likely to contain entities. We explore three different pre-trained language models, namely $\mathrm{BERT_{wwm}}$, $\mathrm{ELECTRA_{base}}$, and $\mathrm{ELECTRA_{legal}}$, as the base encoders and integrate legal features to augment them. For each candidate span, we treat it as a subject to match with potential objects, and then perform max-pooling over the resulting spans to initialize node representations. Considering the unique situations in the judicial field, we design a tailored hypergraph structure. Through high-order reasoning with multiple layers of hypergraph neural networks, entities and relations are finally extracted using a classifier.

The remainder of this paper is organized as follows: Section II surveys the relevant existing literature. Section III presents the overall framework of our model, elaborating on the methods of each module and the rationale behind their design. Section IV details our experimental settings, including extensive comparative and ablation studies to validate our approach. Finally, Section V concludes the paper by summarizing our key contributions and limitations, and discussing potential directions for future research.
The contributions of this study are as follows:
\begin{itemize}
\item We combine the neighborhood-oriented packing strategy with the biaffine mechanism to design a candidate span generator that effectively models entity boundaries in complex texts.
\item We construct a domain-specific dictionary for drug-related legal cases and integrate it as a legal feature to augment the pre-trained encoder.
\item Considering special situations in the judicial field such as joint crimes and combined punishment for multiple crimes, we design an effective hypergraph structure that conforms to judicial practice.
\item Experimental results on the CAIL2022 dataset demonstrate that our proposed method significantly outperforms existing baseline models.
\end{itemize}

\section{Related work}

\subsection{Entity and Relation Extraction}
Early methods for entity and relation extraction often employed traditional neural network-based pipeline approaches. These methods treated named entity recognition and relation extraction as two independent tasks and utilized methods such as Recurrent Neural Networks (RNNs) \cite{hashimoto2013simple}, Convolutional Neural Networks (CNNs) \cite{xu2015semantic}, and Bidirectional Recurrent Convolutional Neural Networks (BRCNNs) \cite{cai2016bidirectional} to model syntactic dependency trees.

Some researchers argue that the pipeline approach neglects the interaction between entities and relations, and advocate for a joint approach to model these interactions. Miwa and Bansal \cite{miwa2016end} were the first to propose an end-to-end neural network model to simultaneously extract entities and relations, achieving a joint representation in a single model through shared parameters. Wang and Lu \cite{wang2020two} proposed a table-sequence encoder, designing two different table and sequence encoders to interact during the representation learning process. Wang et al. \cite{wang2021unire} unified the label spaces of the two sub-tasks into one table and used a unified classifier to predict the label of each cell. Yan et al. \cite{yan2021partition} proposed a partition-filtering network to model the bidirectional interaction between tasks by decomposing feature encoding into partitioning and filtering steps. Kong and Xia \cite{kong2024care} captured the bidirectional interaction between the two sub-tasks through a co-attention network, enabling the model to leverage entity information for relation extraction. Others \cite{sun2019joint,fu2019graphrel,yan2023joint} introduced graph convolutional networks (GCNs) and hypergraph neural networks to jointly learn named entities and relations, first by identifying entity spans and then performing joint reasoning on their types.

In recent years, the emergence of pre-trained language models such as BERT \cite{devlin2018bert} and ELECTRA \cite{clark2020electra} has significantly enhanced the models' ability to understand context, providing new impetus for the pipeline approach. Zhong and Chen \cite{zhong2021frustratingly} proposed a pipeline method constructed on two independent pre-trained encoders, using only the entity model to construct the input for the relation model. Ye et al. \cite{ye2022packed} achieved the state-of-the-art (SOTA) by strategically packing tokens in the pre-trained language model encoder to consider the interrelationships between spans. The debate over whether the pipeline or joint approach is superior remains ongoing, and combining the advantages of both may be a feasible research direction.

\subsection{Information Extraction in the Judicial Field}
With the continuous publicity and transparency of judicial processes, it has become easier to obtain legal documents. As a result, an increasing number of scholars have begun to apply advanced natural language processing technologies in the legal field, proposing many effective methods for legal information extraction and reducing the cost of manual review. For instance, Andrew \cite{andrew2018automatic} designed a system to automatically identify and annotate entities using conditional random fields and regular expressions. Dan et al. \cite{dan2018entity} extracted the relationships between judicial entities in key paragraphs of legal documents through kernel functions and convolutional neural networks (CNNs). More recent works have explored multi-task learning frameworks \cite{chen2020joint}, table-filling approaches \cite{zhang2023joint}, and methods integrating active and transfer learning \cite{zhang2024legalatle}. Despite these advances, most studies involve the direct application of existing technologies, with few fully accounting for the unique characteristics of the judicial domain itself. While some recent efforts, such as the context-aware method proposed by Castano et al. \cite{castano2024enforcing}, have begun to address this gap, there remains a significant need for models that are explicitly designed for the nuances of legal text.

\section{Method}
\renewcommand{\dblfloatpagefraction}{.9}
\begin{figure*}
      \centering
      \includegraphics[width=0.6\linewidth]{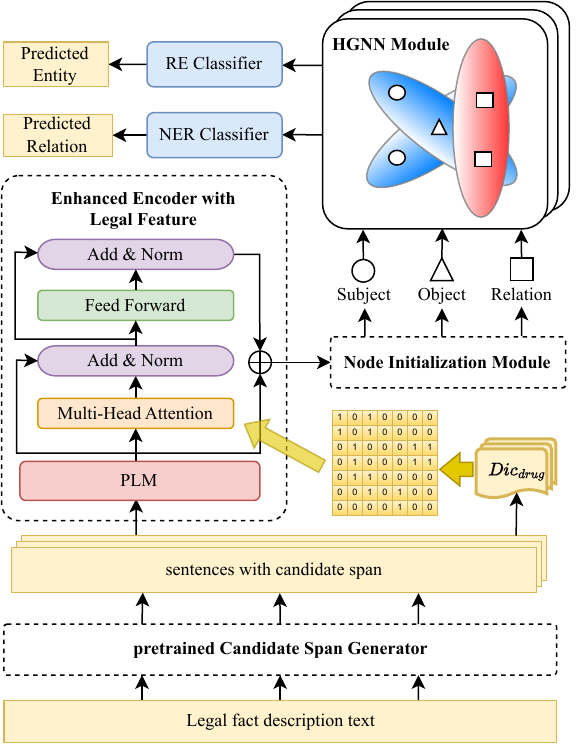}
      \caption{Overview of our model architecture}
      \label{fig2}
\end{figure*}
In this section, we first introduce our candidate span generator, which is trained independently of the other modules. Second, we detail how legal dictionary knowledge is integrated into our Augmented Encoder with Legal Features. Finally, we present the node initialization method and the inference scheme for our custom-designed hypergraph structure. Figure~\ref{fig2} provides an overview of our model architecture.

\subsection{Candidate Span Generator}

\begin{figure}[H]
    \centering
    \includegraphics[width=0.7\linewidth]{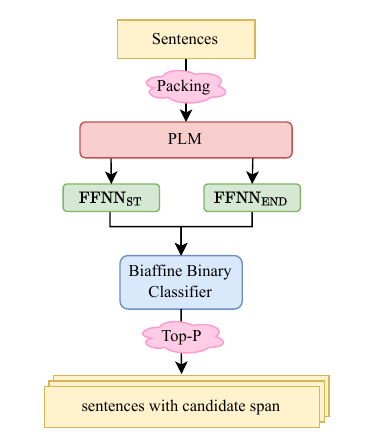}
    \caption{Architecture of the Candidate Span Generator.}
    \label{fig:cs_generator}
\end{figure}

In this module, we independently train a span discriminator to determine whether a given span is likely to contain an entity, thereby generating a set of candidate spans. The architecture is illustrated in Figure~\ref{fig:cs_generator}.

Given an input sentence $X = \{x_1, x_2, \dots, x_n\}$, we employ the neighborhood-oriented packing strategy from \cite{ye2022packed}, which applies levitated markers [O] and [/O]. From the pre-trained language model (PLM), we obtain marker representations $x^{(s)}$ (for [O]) and $x^{(e)}$ (for [/O]). For a span $s_i$ consisting of tokens from $x_{st(i)}$ to $x_{end(i)}$, we compute its start and end representations using two feed-forward networks, $\mathrm{FFNN}_{\mathrm{ST}}$ and $\mathrm{FFNN}_{\mathrm{END}}$:
\begin{equation}
    h_{st}(s_i) = \mathrm{FFNN}_{\mathrm{ST}}([x_{st(i)} \oplus x_{i}^{(s)}])
\end{equation}
\begin{equation}
    h_{end}(s_i) = \mathrm{FFNN}_{\mathrm{END}}([x_{end(i)} \oplus x_{i}^{(e)}])
\end{equation}
where $\oplus$ denotes the concatenation operation. The final span representation $h(s_i)$ is obtained via a biaffine mechanism \cite{yu2020named}:
\begin{equation}
    h(s_i) = [h_{st}(s_i) \oplus 1] U_m [h_{end}(s_i) \oplus 1]^{\top}
\end{equation}
Here, $U_m \in \mathbb{R}^{(d_{st}+1) \times d_u \times (d_{end}+1)}$ is a learnable weight matrix.

To train the discriminator, we use the gold binary tag $y(s_i) \in \{0,1\}$, where $1$ indicates that $s_i$ is a true entity span. The probability $\hat{y}(s_i)$ is calculated using a sigmoid function, and we optimize the binary cross-entropy (BCE) loss, $\mathcal{L}_{span}$:
\begin{equation}
    \hat{y}(s_i) = \mathrm{Sigmoid}(\mathrm{FFNN}(h(s_i)))
\end{equation}
\begin{equation}
    \mathcal{L}_{span} = -[y(s_i)\log\hat{y}(s_i) + (1-y(s_i))\log(1-\hat{y}(s_i))]
\end{equation}

Finally, to generate the candidate span set $S_c$, all possible spans are scored and sorted, and we select the top-$P$ spans. We set $P = \gamma \cdot n$, assuming the number of entity spans is linearly proportional to the sentence length $n$, where $\gamma$ is a hyperparameter.

\subsection{Augmented Encoder with Legal Features}
Following \cite{ye2022packed}, we employ the subject-oriented packing scheme for span pairs to accentuate the subject and object spans within the candidate span set. Given an input sentence $X=\{{{x}_{1}},{{x}_{2}},...,{{x}_{n}}\}$ and a candidate span ${{s}_{i}}\in {{S}_{c}}(X)$, all other spans ${{s}_{j}}\in {{S}_{c}}(X)$ can be the candidate object span using $s_i$ as the subject span. Then insert a pair of solid markers [S] and [/S] before and after the subject span $s_i$ and apply levitated markers $\mathrm{[O]_j}$ and $\mathrm{[/O]_j}$ to each candidate object span $s_j$. The levitated markers are concatenated at the end of the sequence and subsequently fed into the PLM to obtain the general representation $H_G$.

The encoding generated by PLM generally tends to capture the common text representation, but is notably lacking in domain-specific knowledge. To compensate for the deficiency of legal domain information, we integrate legal feature augmentation into the encoder. For the entity relation extraction of Chinese drug-related judgment documents, we construct a dictionary of drug names $\mathrm{Dic}_{\mathrm{drug}}$ as a legal feature. The dictionary is collected by gathering the academic names of various drugs and their common expressions in judgment documents.

Given an input sentence $X=\{{{x}_{1}},{{x}_{2}},...,{{x}_{n}}\}$, we match it against the dictionary $\mathrm{Dic}_{\mathrm{drug}}$ to obtain all possible subsequences that constitute a drug expression. Define $X[a:b]$ as a subsequence of $X$ that starts with $x_a$ and ends with $x_b$. The legal feature of drug names is represented using a mask matrix $M_{\mathrm{dic}}$. $M_{\mathrm{dic}}$ is a $N\times M$ matrix, where the element $m_{ab}$ in row $a$ and column $b$ denotes whether the subsequent $X[a:b]$ is a drug or not.

\begin{equation}
    m_{ab} = \begin{cases}
        1 & \text{if } X[a:b] \in \mathrm{Dic}_{\mathrm{drug}} \\
        0 & \text{otherwise}
    \end{cases}
\end{equation}

We compute a specific legal representation of the input sequence with an extra Transformer encoder layer \cite{vaswani2017attention}. The general representation $H_G$ is projected into different matrices $Q_h$, $K_h$ and $V_h$, where $h$ represents the $h$-th head of multi-head attention. The self-attention function output is calculated from the legal feature mask matrix $M_{\mathrm{dic}}$, denoted as $A_h$.
\begin{equation}
    {{A}_{h}}=\operatorname{Softmax}\left( {M_{\mathrm{dic}}}\And \frac{{{Q}_{h}}K_{h}^{\top}}{\sqrt{{{d}_{k}}}} \right){{V}_{h}}
\end{equation}

where the symbol $\And$ denotes the mask operation.

We concatenate the output of all the attention heads and pass the result through the Feed Forward layer. The specific legal domain representation of the Transformer output integrates the features of the drug dictionary, denoted as $H_L$. Finally, the general representation $H_G$ and the specific legal domain representation $H_L$ are weighted to obtain the Legal feature augmented representation $H_{\mathrm{Legal}}$.
\begin{equation}
    H_{\mathrm{Legal}} = \omega \cdot (\lambda \cdot H_G + (1-\lambda) \cdot H_L)
\end{equation}

where $\omega$ is a fixed hyperparameter for scaling, and $\lambda$ is a learnable scalar weight.

\subsection{Node Initialization Module}

This module generates the initial node representations required for the subsequent Hypergraph Neural Network Module. It initializes three types of nodes—subject, object, and relation—using the legally-augmented representation $H_{\mathrm{Legal}}$.

For each candidate span $s_i=(a,b)$ acting as a subject, its initial representation is derived from the corresponding solid marker pair embeddings:
\begin{equation}
    h_{\mathrm{sub}}(s_i) = \mathrm{FFNN}_{\mathrm{SUB}}([h_{\mathrm{Legal}}^{a-1} \oplus h_{\mathrm{Legal}}^{b+1}])
\end{equation}

For each potential object span $s_j=(c,d)$ relative to the current subject $s_i$, its contextual representation is derived from its levitated markers:
\begin{equation}
    h_{\mathrm{obj}}^{i}(s_j) = \mathrm{FFNN}_{\mathrm{OBJ}}([h_{\mathrm{Legal}}^{c(s)} \oplus h_{\mathrm{Legal}}^{d(e)}])
\end{equation}
where $h_{\mathrm{Legal}}^{c(s)}$ and $h_{\mathrm{Legal}}^{d(e)}$ are the legal augmented representations of the levitated markers inserted for $s_j$.

Similarly, the initial representation for the relation pair $r_{ij}=(s_i,s_j)$ is formed by combining the representations of the subject and its corresponding object:
\begin{equation}
    h_{\mathrm{rel}}(r_{ij}) = \mathrm{FFNN}_{\mathrm{REL}}([h_{\mathrm{sub}}(s_i) \oplus h_{\mathrm{obj}}^{i}(s_j)])
\end{equation}

This process is performed for every candidate span $s_i \in S_c$ serving as a subject. Consequently, while each of the $P$ candidate spans has a single subject representation, each span $s_j$ acquires multiple object representations—one for each of the $P-1$ possible subjects. To resolve this, we apply a max-pooling layer to aggregate these contextual object representations into a single, unique representation for each object span $s_j$:
\begin{equation}
    h_{\mathrm{obj}}(s_j) = \operatorname{MaxPool}_{1 \le i \le P, i \ne j} (h_{\mathrm{obj}}^{i}(s_j))
\end{equation}

\subsection{Hypergraph Neural Network Module}

Information interaction between entities and relations is often beneficial for joint extraction tasks. We exploit this interaction by constructing a hypergraph that models these dependencies. In the legal domain, unique situations such as joint crimes and combined punishment for multiple crimes exist. Therefore, we deliberately incorporate these specific scenarios into our hypergraph design and use a hypergraph neural network to perform higher-order inference.

\subsubsection{Hypergraph Definition}
We construct a hypergraph $G=(V,E)$ with three types of nodes and three types of hyperedges. The node set $V$ comprises candidate subjects $V_{\mathrm{sub}}=\{v_{\mathrm{sub}}^{a} | a\in [1,P]\}$, candidate objects $V_{\mathrm{obj}}=\{v_{\mathrm{obj}}^{b} | b\in [1,P]\}$, and all potential relation pairs $V_{\mathrm{rel}}=\{v_{\mathrm{rel}}^{ab} | a,b\in [1,P], a\ne b\}$. The hyperedge set $E$ contains three subsets, $E_{\mathrm{sor}}$, $E_{\mathrm{jc}}$, and $E_{\mathrm{cp}}$, each designed with a different intent.

\begin{figure}[htbp]
    \centering
    \begin{subfigure}{0.4\linewidth}
		\centering
		\includegraphics[width=0.9\linewidth]{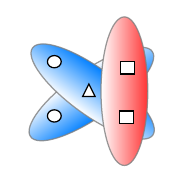}
		\caption{$jc$ hyperedge}
		\label{hyper1}
    \end{subfigure}
    \centering
    \begin{subfigure}{0.4\linewidth}
		\centering
		\includegraphics[width=0.9\linewidth]{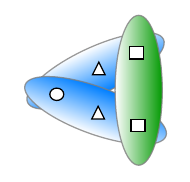}
		\caption{$cp$ hyperedge }
		\label{hyper2}
    \end{subfigure}
    \caption{Examples of custom hyperedges}
    \label{fig:hypergraphs}
\end{figure}

$E_{\mathrm{sor}}$ is designed to model the fundamental subject-object-relation triplet. Each hyperedge $e_{\mathrm{sor}}^{ab} \in E_{\mathrm{sor}}$ connects a subject node $v_{\mathrm{sub}}^{a}$, an object node $v_{\mathrm{obj}}^{b}$, and their corresponding relation node $v_{\mathrm{rel}}^{ab}$.
A joint crime refers to an intentional crime committed by two or more persons jointly. In our task, we model this with $E_{\mathrm{jc}}$ hyperedges, which represent different subjects sharing the same object. Each hyperedge $e_{\mathrm{jc}}^{abC} \in E_{\mathrm{jc}}$ connects two relation nodes, $v_{\mathrm{rel}}^{ac}$ and $v_{\mathrm{rel}}^{bc}$.
Combined punishment for multiple crimes refers to the sentencing of a person who has committed several distinct crimes. We model this with $E_{\mathrm{cp}}$ hyperedges, representing the same subject being involved in relations with different objects. Each hyperedge $e_{\mathrm{cp}}^{Abc} \in E_{\mathrm{cp}}$ connects two relation nodes, $v_{\mathrm{rel}}^{ab}$ and $v_{\mathrm{rel}}^{ac}$. Figure~\ref{fig:hypergraphs} shows examples of these hyperedges.

\subsubsection{Node Representation}
For a subject node $v_{\mathrm{sub}}^a$, an object node $v_{\mathrm{obj}}^b$, and a relation node $v_{\mathrm{rel}}^{ab}$, we use $g^{n}(v_{\mathrm{sub}}^{a})$, $g^{n}(v_{\mathrm{obj}}^{b})$, and $g^{n}(v_{\mathrm{rel}}^{ab})$ to denote their respective output representations from the $n$-th HGNN layer. The initial representations ($n=0$) are provided by the node initialization module: $g^{0}(v_{\mathrm{sub}}^{a}) = h_{\mathrm{sub}}(s_{a})$, $g^{0}(v_{\mathrm{obj}}^{b}) = h_{\mathrm{obj}}(s_{b})$, and $g^{0}(v_{\mathrm{rel}}^{ab}) = h_{\mathrm{rel}}(r_{ab})$.

\subsubsection{Information Incorporation}
Hyperedges act as channels for information propagation. Since hyperedges do not have their own representations, we compute their messages by aggregating the representations of their constituent nodes.

For a $sor$ hyperedge $e_{\mathrm{sor}}^{ab}$, its message incorporates information from the three connected nodes at the $(n-1)$-th layer:
\begin{equation}
   {{h}^{n}}(e_{\mathrm{sor}}^{ab})={{g}^{n-1}}(v_{\mathrm{sub}}^{a})\oplus {{g}^{n-1}}(v_{\mathrm{obj}}^{b})\oplus {{g}^{n-1}}(v_{\mathrm{rel}}^{ab})
\end{equation}
\begin{equation}
    {{m}^{n}}(e_{\mathrm{sor}}^{ab})={{\mathrm{FFNN}}_{\mathrm{SOR}}}({{h}^{n}}(e_{\mathrm{sor}}^{ab}))
\end{equation}

For a $jc$ hyperedge $e_{\mathrm{jc}}^{abC}$, its message aggregates information from two relation nodes sharing the same object:
\begin{equation}
    m^{n}(e_{\mathrm{jc}}^{abC}) = \mathrm{FFNN}_{\mathrm{JC}}([g^{n-1}(v_{\mathrm{rel}}^{aC}) \oplus g^{n-1}(v_{\mathrm{rel}}^{bC})])
\end{equation}

For a $cp$ hyperedge $e_{\mathrm{cp}}^{Abc}$, its message aggregates information from two relation nodes sharing the same subject:
\begin{equation}
    m^{n}(e_{\mathrm{cp}}^{Abc}) = \mathrm{FFNN}_{\mathrm{CP}}([g^{n-1}(v_{\mathrm{rel}}^{Ab}) \oplus g^{n-1}(v_{\mathrm{rel}}^{Ac})])
\end{equation}

\subsubsection{Update Mechanism}
For each node $v \in V$, messages from all its adjacent hyperedges $E_v$ are aggregated via an attention mechanism. This aggregated message is then used to update the node's representation in a residual manner. The normalized attention weight $\beta_e^n$ for each edge $e \in E_v$ is calculated as:
\begin{equation}
    \alpha_{e}^{n} = w^{\top}\phi(W[\mathrm{FFNN}_{V}(g^{n-1}(v)) \oplus m^{n}(e)]+b)+c
\end{equation}
\begin{equation}
    \beta_{e}^{n} = \frac{\exp(\alpha_{e}^{n})}{\sum_{e' \in E_{v}} \exp(\alpha_{e'}^{n})}
\end{equation}
Finally, the node representation is updated:
\begin{equation}
    g^{n}(v) = g^{n-1}(v) + \sum_{e \in E_{v}} \beta_{e}^{n} m^{n}(e)
\end{equation}
where $\phi(\cdot)$ is a non-linear activation function, and $w, W, b, c$ are trainable parameters. Entity nodes (subjects and objects) only receive messages from $sor$ hyperedges, while relation nodes can receive messages from all three types of hyperedges.

\subsection{Loss Function}
After $N$ iterations of the HGNN, we use the final node representations $g^N(v)$ for classification. For each candidate entity span $s_i \in S_c(X)$, we concatenate its final subject and object representations and feed them into an NER classifier to compute the probability distribution over the entity types $\{T_{\mathrm{entity}}\} \cup \{\text{null}\}$:
\begin{equation}
    P_e(\hat{y}_e | s_i) = \mathrm{softmax}(\mathrm{FFNN}_{E}([g^{N}(v_{\mathrm{sub}}^{i}) \oplus g^{N}(v_{\mathrm{obj}}^{i})]))
\end{equation}

For each span pair $r_{ij}=(s_i,s_j)$, we feed the final representation of the corresponding relation node into an RE classifier to compute the probability distribution over the relation types $\{T_{\mathrm{relation}}\} \cup \{\text{null}\}$:
\begin{equation}
    P_r(\hat{y}_r | r_{ij}) = \mathrm{softmax}(\mathrm{FFNN}_{R}(g^{N}(v_{\mathrm{rel}}^{ij})))
\end{equation}

We use the cross-entropy loss for both the NER and RE tasks. The total loss $\mathcal{L}$ is the sum of the two:
\begin{equation}
    \mathcal{L}_{\mathrm{ner}} = -\sum_{s_i \in S_c} \log(P(\hat{y}_e = y_e | s_i))
\end{equation}
\begin{equation}
    \mathcal{L}_{\mathrm{re}} = -\sum_{s_i, s_j \in S_c, i \ne j} \log(P(\hat{y}_r = y_r | r_{ij}))
\end{equation}
where $y_e$ and $y_r$ are the ground-truth labels. The final loss is $\mathcal{L} = \mathcal{L}_{\mathrm{ner}} + \mathcal{L}_{\mathrm{re}}$.

\section{Experiments and Results}
\subsection{Experimental Setup}

\subsubsection{Dataset}
We conduct experiments on the dataset from the Challenge of AI in Law (CAIL) 2022 Information Extraction Competition\footnote{http://cail.cipsc.org.cn}. The dataset is derived from legal documents of Chinese drug-related criminal cases published online. It focuses on the task of extracting entities and relations from texts concerning three representative crimes: drug trafficking, illegal possession of drugs, and providing venues for drug users. The dataset contains 1,763 documents with factual descriptions, annotated with 5 predefined entity types and 4 relation types. To facilitate model training and validation, we split the official training set into a new training set and a validation set in a 3:1 ratio, while holding out the original test set for final evaluation.

\subsubsection{Evaluation Metrics} 
For both subtasks, we evaluate the performance using Precision, Recall, and F1 scores with micro-averaging. For NER, a correctly extracted result requires the model to accurately predict both the boundaries and types of the entities. For RE, it requires the model to correctly predict both the boundaries and types of both the subject and object entities, as well as the relation between them.
\begin{equation}
    \mathrm{Precision} = \frac{\mathrm{correct}_{\mathrm{num}}}{\mathrm{Predict}_{\mathrm{num}}}
\end{equation}
\begin{equation}
    \mathrm{Recall} = \frac{\mathrm{correct}_{\mathrm{num}}}{\mathrm{True}_{\mathrm{num}}}
\end{equation}
\begin{equation}
    \mathrm{F1} = 2 \cdot \frac{\mathrm{Precision} \cdot \mathrm{Recall}}{\mathrm{Precision} + \mathrm{Recall}}
\end{equation}

\subsubsection{Implementation Details} 

Due to the particularity of the Chinese language, we adopt Chinese pre-trained language models as encoders, including \textit{chinese-bert-wwm-ext}, \textit{chinese-electra-base}, and \textit{legal-electra-base} \cite{cui2021pre}. Additionally, the \textit{legal-electra-base} model is specifically oriented towards the judicial field, having been pretrained on large-scale Chinese legal document data. We employ the same cross-sentence information incorporation method as that utilized in \cite{zhong2021frustratingly,ye2022packed}, which extends the original sentence to a fixed window size of 512 by including its left and right context. The Top-P ratio $\gamma$ is set to 0.5. The initial value for $\omega$ is 0.5, while $\lambda$ is set to 1.0. The output sizes of all FFNN layers are tuned within the range [256, 400, 512, 768]. We train our model with the Adam optimizer and a linear scheduler with a warmup ratio of 0.1 and an Adam epsilon of $1 \times 10^{-8}$. The learning rate for the PLM is tuned within [$2 \times 10^{-5}, 3 \times 10^{-5}, 5 \times 10^{-5}$], while for other modules, the learning rate is tuned within [$5 \times 10^{-5}, 1 \times 10^{-4}, 2 \times 10^{-4}$]. All experiments are conducted on Nvidia V100 GPUs.

\renewcommand{\dblfloatpagefraction}{.9}
\begin{table*}[htbp]
\renewcommand\arraystretch{1.25}
\caption{\textbf{Overall Evaluation}}
\begin{center}
\begin{tabular}{ccccccccccccc} 
\toprule 
\multicolumn{3}{c}{\multirow{2}*{Encoder}} & \multicolumn{3}{c}{\multirow{2}*{Model}} & \multicolumn{3}{c}{Entity}&\multicolumn{3}{c}{Relation}\\
\multicolumn{3}{c}{}&\multicolumn{3}{c}{}&Precision&Recall&F1&Precision&Recall&F1&\\  
\hline 
\multicolumn{3}{c}{\multirow{6}*{$\mathrm{BERT_{wwm}}$}}&\multicolumn{3}{c}{UniRE\cite{wang2021unire}}&86.7&\textbf{89.3}&88.0&81.8&78.2&80.0\\   
\multicolumn{3}{c}{}&\multicolumn{3}{c}{PFN-nested\cite{yan2021partition}}&85.0&66.8&74.8&\textbf{87.4}&68.1&76.5\\   
\multicolumn{3}{c}{}&\multicolumn{3}{c}{CARE\cite{kong2024care}}&84.2&73.1&78.2&81.8&75.1&78.3\\   
\multicolumn{3}{c}{}&\multicolumn{3}{c}{PURE(Full)\cite{zhong2021frustratingly}}&89.6&88.1&88.9&76.7&79.0&77.8\\   
\multicolumn{3}{c}{}&\multicolumn{3}{c}{PL-Marker\cite{ye2022packed}}&92.1&79.3&85.2&80.3&\textbf{81.8}&81.0\\   
\multicolumn{3}{c}{}&\multicolumn{3}{c}{Legal-KAHRE(ours)}&\textbf{93.4}&89.2&\textbf{91.2}&85.2&79.8&\textbf{82.4}\\   
\hline 
\multicolumn{3}{c}{\multirow{6}*{$\mathrm{ELECTRA_{base}}$}}&\multicolumn{3}{c}{UniRE\cite{wang2021unire}}&87.2&\textbf{90.0}&88.6&82.4&81.4&81.9\\   

\multicolumn{3}{c}{}&\multicolumn{3}{c}{PFN-nested\cite{yan2021partition}}&85.7&65.9&74.5&81.1&71.7&76.1\\   
\multicolumn{3}{c}{}&\multicolumn{3}{c}{CARE\cite{kong2024care}}&86.5&72.1&78.6&82.9&75.8&79.2\\   
\multicolumn{3}{c}{}&\multicolumn{3}{c}{PURE(Full)\cite{zhong2021frustratingly}}&89.8&88.7&89.3&74.4&77.5&75.9\\   
\multicolumn{3}{c}{}&\multicolumn{3}{c}{PL-Marker\cite{ye2022packed}}&92.8&79.4&85.6&82.0&\textbf{82.8}&82.4\\   
\multicolumn{3}{c}{}&\multicolumn{3}{c}{Legal-KAHRE(ours)}&\textbf{93.7}&89.1&\textbf{91.4}&\textbf{83.9}&82.6&\textbf{83.3}\\   
\hline 
\multicolumn{3}{c}{\multirow{6}*{$\mathrm{ELECTRA_{legal}}$}}&\multicolumn{3}{c}{UniRE\cite{wang2021unire}}&87.2&\textbf{89.2}&88.2&83.1&80.7&81.9\\   
\multicolumn{3}{c}{}&\multicolumn{3}{c}{PFN-nested\cite{yan2021partition}}&85.7&66.5&74.9&85.0&72.3&78.2\\   
\multicolumn{3}{c}{}&\multicolumn{3}{c}{CARE\cite{kong2024care}}&85.0&73.3&78.7&85.1&74.1&79.2\\   
\multicolumn{3}{c}{}&\multicolumn{3}{c}{PURE(Full)\cite{zhong2021frustratingly}}&90.4&87.4&88.9&76.3&81.6&78.9\\   
\multicolumn{3}{c}{}&\multicolumn{3}{c}{PL-Marker\cite{ye2022packed}}&92.9&78.8&85.3&82.5&\textbf{83.1}&82.8\\   
\multicolumn{3}{c}{}&\multicolumn{3}{c}{Legal-KAHRE(ours)}&\textbf{94.5}&89.0&\textbf{91.7}&\textbf{87.1}&80.8&\textbf{83.8}\\   
\bottomrule 
\end{tabular}
\end{center}
\label{tab1}
\end{table*}

\subsection{Comparative Analysis with Baselines}

We compare our model with several strong baseline models on the CAIL2022 dataset, using three different pre-trained encoders. The baseline models include:
\begin{itemize}
    \item \textbf{UniRE} \cite{wang2021unire}: Uses a table containing all word pairs from a sentence and applies a unified classifier to predict each cell's label.
    \item \textbf{PFN} \cite{yan2021partition}: Constructs a partition filter network to model bidirectional interactions between the NER and RE tasks. Following the authors' recommendations for Chinese datasets, we use the \textit{PFN-nested} variant, which is better at leveraging entity tail information.
    \item \textbf{CARE} \cite{kong2024care}: Adopts a parallel encoding strategy to learn separate representations for each task and uses a co-attention module to capture their interaction.
    \item \textbf{PURE} \cite{zhong2021frustratingly}: A pipeline method where the input for the relation model consists only of the output from the entity model. To prioritize performance, we use the \textit{PURE(Full)} model rather than the \textit{PURE(Approx)} version.
    \item \textbf{PL-Marker} \cite{ye2022packed}: Designs packing strategies for NER and RE to enhance the encoder, thereby considering the interrelations between spans.
\end{itemize}

We present the experiment results of our model (Legal-KAHRE) and compare them with several strong methods, as shown in TABLE~\ref{tab1}. It should be noted that the results of the Legal-KAHRE shown in the table adopt the hypergraph with the $sorjccp$ structure, because it achieves a balance in the effects of named entity recognition and relation extraction. Under the three pre-trained encoders, Legal-KAHRE has achieved leading F1 scores in both entity recognition and relation extraction. Specifically, it improves by at least 2.1 in NER (compared with PURE) and at least 0.9 in RE (compared with PL-Marker). Legal-KAHRE has achieved the best scores under the encoder $\mathrm{ELECTRA_{legal}}$ both on NER (91.7 vs. 91.4 vs. 91.2) and RE (83.8 vs. 83.3 vs. 82.4). This may be attributed to the fact that $\mathrm{ELECTRA_{legal}}$ was pre-trained on judicial documents during the pre-training stage and is thus more suitable for tasks in the judicial field. It is worth noting that the impact of different encoders on different models may vary, such as UNiRE (88.2 vs. 88.6) and PFN-nested (74.5 vs. 74.8, 76.1 vs. 76.5). Overall, Legal-KAHRE demonstrates a significant improvement over the previous baselines.

\subsection{Ablation Study}
We evaluated the roles of different components in our model, including the candidate span generator, legal feature augmentation, and hypergraph neural network modules. The results under the $\mathrm{BERT_{wwm}}$ encoder are presented in Table~\ref{tab2}. First, we replaced the candidate span generator with a structure identical to the NER module of PL-Marker but used a binary loss to generate candidate spans. Without the candidate span generator, the NER performance of the model dropped by 2.7 points, demonstrating the critical role of the candidate span generator in NER. However, due to the gains from the subsequent modules, its NER and RE performances are still far superior to those of PL-Marker. Next, we removed the legal dictionary augmentation in the encoder and only retained the original pre-trained encoder. After removing legal feature augmentation, the RE performance of the model decreased by 1.8, indicating that the introduction of the legal dictionary is beneficial to the RE task. Finally, we completely removed the hypergraph neural network module and directly output the results of the node initialization module through the classifier. The model lacking the high-order reasoning of the hypergraph experienced a decrease of 0.8 in NER and 1.7 in RE. This shows that the hypergraph neural network module establishes beneficial information interactions between entities and relations and effectively models the unique characteristics in the judicial domain.

\begin{table}[H]
\renewcommand\arraystretch{1.25}
\caption{\textbf{Ablation Study}}
\centering
\begin{tabular}{ccccccccccc} 
\toprule 
\multicolumn{2}{c}{Setting}&Entity&Relation\\  
\hline 
\multicolumn{2}{c}{Default}&91.2&82.4\\   
\multicolumn{2}{c}{- Candidate Span Generator}&88.5&82.4\\
\multicolumn{2}{c}{- Legal Feature}&91.1&80.6\\
\multicolumn{2}{c}{- HGNN Module}&90.4&80.7\\
\bottomrule 
\end{tabular}
\label{tab2}
\end{table}

\subsection{Effect of Hypergraph Structure}
We investigated the impacts of different hypergraph settings, including the topological structures of hypergraphs and the number of layers of hypergraph neural networks.

Table~\ref{tab3} presents the experimental results for hypergraph settings with different topological structures under the $\mathrm{BERT_{wwm}}$ encoder. The names of the hypergraphs represent the corresponding hyperedges used, and $no edge$ indicates that no hypergraph structure was employed. The experimental results reveal that several hypergraph topologies containing the $sor$ hyperedge perform well in the NER task, all achieving F1 scores exceeding 91. In sharp contrast, the three topologies of $jc$, $cp$, and $jccp$ do not contain the $sor$ hyperedge. Consequently, these topologies lack edges connected to entity nodes, meaning that they cannot optimize the representation of entities and may even negatively impact entity recognition performance, falling below the 90.4 achieved by using only the node initialization module. The $cp$ topology obtains the highest F1 score of 82.8 in the RE task, and the hypergraph topologies containing either the $jc$ hyperedge or the $cp$ hyperedge outperform $no edge$ in relation extraction. This highlights the benefits of the $jc$ and $cp$ hyperedges for relation extraction and also verifies that the portrayal of these two hyperedges for domain-specific situations in the legal field, such as joint crimes and combined punishment for multiple crimes, is effective. Overall, the $sor$ hyperedge is highly advantageous for entity recognition, while the $jc$ and $cp$ hyperedges are more proficient in relation extraction. The $sorjccp$ topology, which combines the two, becomes a preferable choice as it ensures the effect of entity recognition and also has a superior effect in relation extraction. Meanwhile, it demonstrates the advantages of information interaction between entities and relations.

\begin{table}[!htbp]
\renewcommand\arraystretch{1.25}
\caption{\textbf{Effect of different graph topologies}}
\centering
\begin{tabular}{ccccccccccc} 
\toprule 
\multicolumn{2}{c}{Hypergraph}&Entity&Relation\\  
\hline 
\multicolumn{2}{c}{$no edge$}&90.4&80.7\\   
\multicolumn{2}{c}{$sor$}&91.2&80.2\\
\multicolumn{2}{c}{$jc$}&90.0&81.9\\
\multicolumn{2}{c}{$cp$}&90.2&82.8\\
\multicolumn{2}{c}{$sorjc$}&91.1&81.4\\
\multicolumn{2}{c}{$sorcp$}&91.2&81.7\\
\multicolumn{2}{c}{$jccp$}&90.1&82.4\\
\multicolumn{2}{c}{$sorjccp$}&91.2&82.4\\
\bottomrule 
\end{tabular}
\label{tab3}
\end{table}

\begin{figure}[htbp]
    \centering
    \begin{subfigure}{0.49\linewidth}
		\centering
		\includegraphics[width=0.9\linewidth]{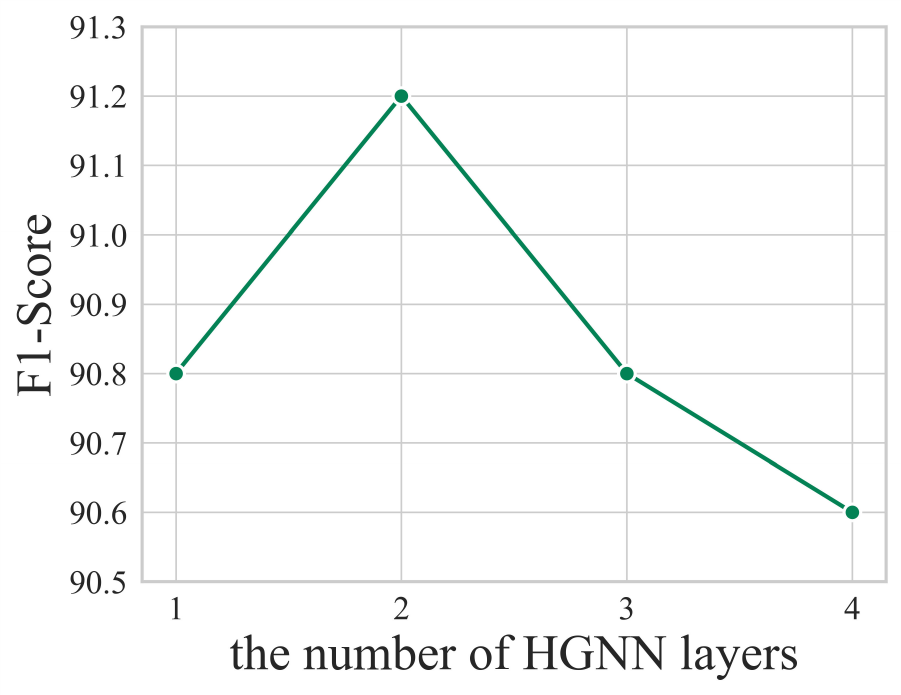}
		\caption{NER F1}
    \end{subfigure}
    \centering
    \begin{subfigure}{0.49\linewidth}
		\centering
		\includegraphics[width=0.9\linewidth]{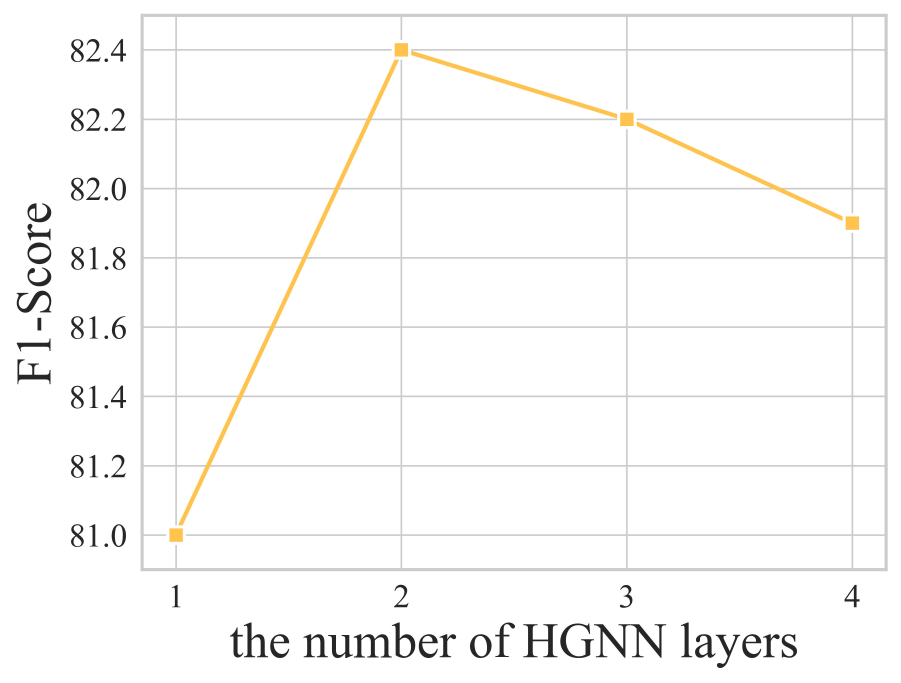}
		\caption{RE F1 }
    \end{subfigure}
    \caption{Effect of numbers of HGNN layers}
    \label{hgnn}
\end{figure}

Second, we analyzed the impact of the number of HGNN layers on model performance. As illustrated in Figure~\ref{hgnn}, using two layers of HGNN yields the best results for both NER and relation extraction. Employing more layers leads to a decline in performance, which may be attributed to the over-smoothing problem common in graph neural networks \cite{rusch2023survey}.


\section{Conclusion}
In this paper, we propose an entity and relation extraction method based on knowledge augmentation and hypergraph neural networks, specifically designed for legal documents. We integrate the neighborhood-oriented packing strategy with the biaffine mechanism to design a candidate span generator, which effectively models entity boundaries in complex legal texts. Additionally, we collect domain-specific knowledge from the legal field to construct a dictionary, which augments the encoder's features and enhances its adaptability to the judicial domain. To address unique scenarios in the judicial field, such as joint crimes and combined punishment for multiple crimes, we design a specialized hypergraph structure to capture these domain-specific characteristics. Experimental results demonstrate the effectiveness of both encoder augmentation and the tailored hypergraph structure for the judicial domain, as well as the superiority of our proposed method over existing approaches. In future work, we will continue to explore methods applicable to a broader range of judicial domains, beyond the drug-related field.

\section*{Acknowledgment}

This research was supported by the National Natural Science Foundation of China (NSFC) under Grant 72071029 and 72231010.

\bibliographystyle{IEEEtran}
\bibliography{refs}

\end{document}